\RequirePackage{fix-cm}
\documentclass{svjour3}                     % onecolumn (standard format)
\smartqed  % flush right qed marks, e.g. at end of proof
\usepackage{amssymb}
\usepackage{tabularx} 
\usepackage{graphicx}
\usepackage{algorithm, algorithmic}
\usepackage{amsmath}
\usepackage{color}
\usepackage{multirow}
\begin{document}

\title{Video and Accelerometer-Based Motion Analysis for Automated Surgical Skills Assessment}


\author{Aneeq Zia \and Yachna Sharma \and Vinay Bettadapura \and Eric L. Sarin \and Irfan Essa %etc.
}


\institute{Aneeq Zia \at
              College of Computing, Georgia Tech \\
              Tel.: +1-404-924-0313\\
              \email{aneeqzia@gmail.com}           %  \\
}

\date{Received: date / Accepted: date}

\maketitle
\begin{abstract}
	\textit{Purpose}: 
	Basic surgical skills of suturing and knot tying are an essential part of medical training. Having an automated system for surgical skills assessment could help save experts time and improve training efficiency. There have been some recent attempts at automated surgical skills assessment using either video analysis or acceleration data. In this paper, we present a novel approach for automated assessment of OSATS based surgical skills and provide an analysis of different features on multi-modal data (video and accelerometer data).	
	
\noindent \textit{Methods}: 
	We conduct the largest study, to the best of our knowledge, for basic surgical skills assessment on a dataset that contained video and accelerometer data for suturing and knot-tying tasks. We introduce ``entropy based'' features -- Approximate Entropy (‘ApEn’) and Cross-Approximate Entropy (‘XApEn’), which quantify the amount of predictability and regularity of fluctuations in time-series data. The proposed features are compared to existing methods of Sequential Motion Texture (SMT), Discrete Cosine Transform (DCT) and Discrete Fourier Transform (DFT), for surgical skills assessment.
		
\noindent \textit{Results}: We report average performance of different features across all applicable OSATS criteria for suturing and knot tying tasks. Our analysis shows that the proposed entropy based features out-perform previous state-of-the-art methods using video data. For accelerometer data, our method performs better for suturing only. We also show that fusion of video and acceleration features can improve overall performance with the proposed entropy features achieving highest accuracy.  
	
\noindent \textit{Conclusions}: Automated surgical skills assessment can be achieved with high accuracy using the proposed entropy features. Such a system can significantly improve the efficiency of surgical training in medical schools and teaching hospitals.
	
%	We present a novel approach for automated assessment of OSATS based surgical skills from videos and accelerometer data. We introduce entropy based features, which quantify the amount of predictability and regularity of fluctuations in time-series data. We compare our approach to existing methods for doing surgical assessment and our results provide valuable insights on features and modalities to express a surgical trainee's skills. Our experiments show that motion features from videos provide better assessments as compared to acceleration data due to wholesomeness of motion information in the videos. We also note that the proposed entropy based features outperform existing state-of-the-art features described in the literature when tested over a large data set of surgical skills with varying expertise levels.
	
	\keywords{Surgical skills assessment \and Computer vision \and Machine learning \and Multi-modal data}
	
	%Assessment of surgical skills using videos has gained popularity in recent years. However, there exist few studies that have reported results on a large number of participants. While some works have reported assessments using sensor data collected via accelerometers, very few of them have reported comparison of video and acceleration based assessment of surgical skills. In this work, we present the largest study of its kind to assess OSATS based surgical skills from videos and accelerometer data and compare several existing features for both video and acceleration data. Our results provide interesting insights on features and modalities to express a surgical trainee's skills in terms of motion data. We conclude that motion features from videos provide better assessments as compared to acceleration data due to wholesomeness of motion information in the videos. We also note that the proposed entropy based features outperform existing state-of-the-art features described in the literature. 
\end{abstract}
%\vspace{-5pt}
\section{Introduction}
%\vspace{-5pt}
\label{sec:introduction}
Surgical trainees are required to acquire specific skills during the course of their residency before performing real surgeries. Surgical training involves constant practice of skills and seeking feedback from supervising surgeons, who are generally very busy. Furthermore, manual assessments, even by experts are subjective and prone to errors. Objective Structured Assessment of Technical Skills (OSATS) is adopted in most medical schools as a standard to assess surgical residents~\cite{28martin1997objective}. The OSATS grading scheme includes specific criteria like Respect for Tissue (RT), Time and Motion (TM), Instrument Handling (IH), Suture Handling (SH), Flow of Operation (FO), Knowledge of Procedure (KP), and Overall Performance (OP). While adopting OSATS grading system reduces the subjectivity of assessment to some extent, it is quite resource limiting as only a few expert surgeons can do the scoring and provide feedback. 

To address the time consuming and subjective nature of manual assessments, recent works have proposed techniques that analyze motion from videos~\cite{bettadapura2013augmenting,isbi,sharmavideo,zia2015automated} and wearable sensors to assess surgical skills~\cite{13ahmidi2012objective,trejos2008design}. These approaches showcase different feature types to perform OSATS assessments. We propose entropy based features that quantify the amount of predictability and regularity of fluctuations in time-series data inherent in surgical motions. We show, using experiments on a large data set, that these new features outperform existing features types for surgical skills assessment. Additionally, we also extend our comparison to include different feature types for both acceleration data (from wearable sensors) and video analysis. 

%To the best of our knowledge, there have been no substantial efforts to compare the automated OSATS assessments using video and accelerometer data. However, a comparison of different features types and data modalites (video, acceleration) is lacking in literature. 

\emph{\textbf{Contributions:}} (1)  We propose a novel way of leveraging the irregularity in surgeon motions to assess surgical skills using entropy based features.  (2) We provide a comparison of existing techniques on both video and acceleration data. (3) We perform the biggest study, to our knowledge, on assessing basic surgical tasks like suturing and knot tying using video and acceleration data.
%\vspace{-5pt}

%\vspace{-2pt}
\section{Background}
%\textbf{*some parts taken from miccai 2014 paper*}
%\vspace{-6pt}
The problem of automated surgical skills assessment has recently seen some good progress. Pioneering efforts were based on robotic minimally invasive surgery (RMIS) and focused on gesture recognition and skill assessment using Hidden Markov Models \cite{rosen2001markov,09reiley2009decomposition}. Some other methods like linear dynamical systems (LDS) and bag of words (BoW) models have also been used for RMIS based skill assessments \cite{haro2012surgical,zappella2013surgical}.

Video based skill assessments have also gained interest in recent years. For example, Augmented BoW (A-BoW) features introduced in~\cite{bettadapura2013augmenting}, modeled motion as short sequences of events and the underlying temporal and structural information is automatically discovered and encoded into BoW models. Other techniques based on the holistic analysis of time series data include Motion Texture(MT)~\cite{isbi} for prediction of surgical skill scores by encoding video motion dynamics into frame kernel matrices followed by texture analysis. Sequential Motion Textures (SMT) was proposed in \cite{sharmavideo} which included the sequential information into MT technique by dividing the time series into sequential time windows. More recently frequency based features (DFT and DCT)~\cite{zia2015automated,zia2016automated} have also been used for surgical skill classification. An exhaustive analysis of video based OSATS assessments is presented in~\cite{zia2016automated}, however, results for only video data are presented.

The techniques mentioned above do provide encouraging results for video based surgical skill assessment.  However, these studies use very few participants which limits their ability to capture the wide variation in surgical skills. An expert surgeon's hand motion might be more clean, distinct, ordered and sequential as compared to a non-expert and having more samples helps capture skills of varying levels. Most of the works mentioned above have focused on granularity (MT, SMT) and repetitiveness (DFT, DCT) of motion, however, disorder in motion has not been addressed.  Also, they do not include studies on  wearable motion sensing devices such as accelerometers that may provide precise motion information for surgical skills assessment. 

In the computer vision literature, there has been some recent progress in assessing quality of actions, especially in the sports domain. In \cite{pirsiavash2014assessing}, the authors presented an approach of using pose with frequency features to predict sports scores. More recently, \cite{venkataraman2015dynamical} used entropy features with pose to predict scores for Olympic diving videos. We take inspiration from their work and propose to encode predictability in surgical motions via entropy based features for skills assessment.
%In addition, there is a lack of comparative analysis using other modalities such as acceleration. 

In this work, we provide comparative analysis of several features using video and acceleration data on the largest group of participants, to the best of our knowledge, published thus far. We also propose entropy based features (encoding orderliness in motion) and demonstrate their efficacy as they outperform other types of features for both acceleration and video data.

%\vspace{-5pt}
\section{Methodology}
%\vspace{-6pt}
We believe that the difference in the predictability of the motions of surgeons with varying skills levels can be used to assess the basic surgical skills, for specific tasks like suturing and knot tying. An expert will have more predictable hand motion while a beginner will exhibit erratic and irregular patterns. We propose to measure this difference in predictability of motions using entropy based features \textit{``Approximate Entropy (ApEn)"} and \textit{``Cross Approximate Entropy (XApEn)"}. 

\begin{figure}[t]
	\centering
	\includegraphics[width=1.0\columnwidth]{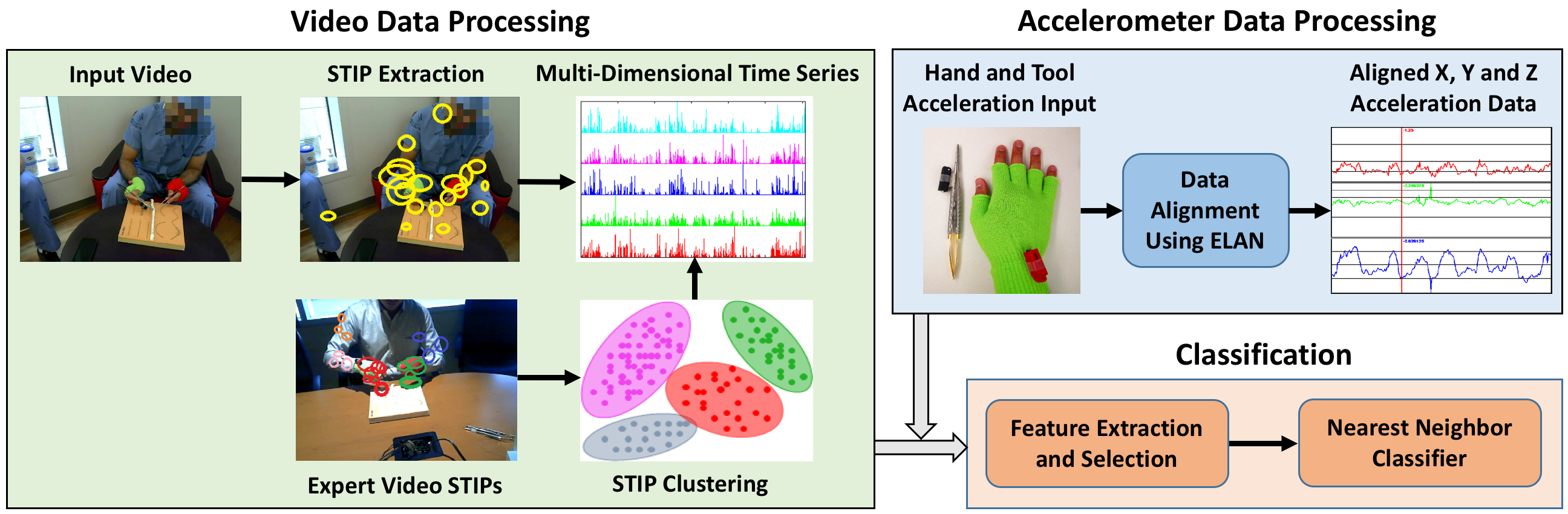}
	\caption{Flow diagram for processing the video and accelerometer data.}
	\label{fig:flow_diagram}
%	\vspace{-2pt}
\end{figure}

Figure \ref{fig:flow_diagram} shows the flow diagram for video and accelerometer data processing. For videos, we follow the standard approach, as used by~\cite{sharmavideo,zia2015automated,zia2016automated}, for encoding motion information from video data into a multi-dimensional time series using Spatio-Temporal Interest Points (STIPS). As presented in these previous works, we use expert videos to learn motion classes via $k$-means clustering for different number of clusters $K$. These motion classes are then used to convert each video into a multi-dimensional time series $T_v \in \Re^{K \times N}$, where $K$ represents the number of motion classes learned (number of clusters used in $k$-means clustering) and $N$ is the number of frames in the video. For the accelerometer data, the $x$, $y$ and $z$ acceleration time series captured from two accelerometers for each surgical task are concatenated to produce a time series $T_a \in \Re^{6 \times Q}$, where $Q$ is the number of samples captured. We also use individual accelerometer time series data for our analysis as discussed in Section 5. The time series data obtained for both modalities is then used for feature extraction and skill prediction. Sequential forward feature selection (SFFS) is used to reduce the dimensionality of the features used in comparison and a Nearest-Neighbor (NN) classifier is used for classification. 

%We compare entropy based features with previous state-of-the-art features such as Sequential Motion Texture (SMT), Discrete Fourier Transform (DFT) and Discrete Cosine Transform (DCT). We do not use the traditional Hidden Markov Model (HMM) and Bag-of-Words (BoW and A-BoW) in our comparison as these have already been shown to perform poorly in \cite{zia2015automated} for this domain.
%\vspace{-5pt}
\subsection{Entropy Features for Skill Assessment}
%\vspace{-6pt}
Entropy is a measure of uncertainty in any data. 
%It has been widely used in various fields for different applications and studies including machine learning e.g feature selection. 
For time series data analysis, entropy based features are used to quantify the amount of predictability and regularity of fluctuations in the time series data. In this paper, we propose to use entropy based features \textit{`Approximate Entropy'} and \textit{`Cross Approximate Entropy'} for assessing the skill of surgeons on various OSATS criteria. The details of both these techniques are given below.

\noindent\emph{\textbf{Approximate Entropy:}} 
Approximate entropy is a measure of regularity in time series data initially proposed in \cite{pincus1991approximate}. A more predictable time series would have a low approximate entropy value whereas an irregular time series would have a higher entropy. For a one-dimensional time series, the approximate entropy \textit{`ApEn'} is dependent on three parameters: embedding dimension ($m$), radius ($r$) and time delay ($\tau$). The embedding dimension ($m$) represents the length of the series which is being checked for repeatability, the radius ($r$) is used for local probabilities estimation and time delay ($\tau$) is selected in a way to make the components of the embedding vector independent enough. For a given time series $T \in \Re^N$, we form a sequence of embedding vectors $x(1), x(2), \ldots, x(N-m+1)$, where $x(i)$ is given by $x(i) = [T_i,T_{i+\tau},\ldots,T_{i+(m-1)\tau}]$, for $ 1\leq i\leq N-(m-1)\tau$. Then, for each embedding vector $x(i)$, the frequency of repeatable patterns $C^{m}_i(r)$ is calculated by
\begin{equation}
	C^{m}_i(r) = \frac{1}{N-(m-1)\tau}\sum_{j}H(r-dist(x(i),x(j)))
\end{equation}
where $H$ is the Heaviside step functions and $dist(x(i),x(j)) = \max(|T(i+(k-1)\tau)-x(j+(k-1)\tau)|)$ for $k \in [1,2,\ldots,m]$. The conditional frequency estimates are calculated by
\begin{equation}
	\Omega^m(r) = \frac{1}{N-(m-1)\tau}\sum_{i=1}^{N-(m-1)\tau}ln(C^m_i(r)
\end{equation}
$\Omega(r)$ is then used to calculate the approximate entropy for the time series $T \in \Re^N$ as $ApEn(m,r,\tau) = \Omega^m(r) - \Omega^{m+1}(r).$

\begin{figure}[b]
	\centering
	\includegraphics[width=1.0\columnwidth]{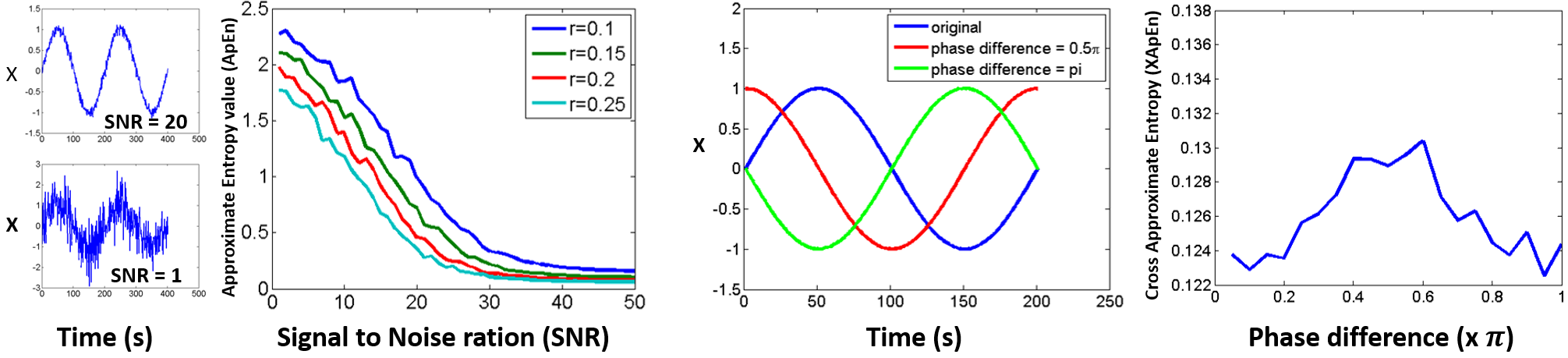}
	\caption{(a) Sample sine waves with different SNR. (b) Variation of \textit{`ApEn'} with respect to SNR (c) Sample sine waves with different phases (d) Variation of \textit{`XApEn'} with respect to phase difference between signals }
	\label{fig:toy}
	%	\vspace{-2pt}
\end{figure}

In order to show how \textit{`ApEn'} varies for signals with different predictability, we generate a set of sinusoids $V$. A pure sine wave without any noise can be considered as completely `predictable' since it has a fixed repeating pattern. However, adding noise to the same function would make it less predictable. We induce white Gaussian noise into our set of sinusoids $V$ to vary the signal-to-noise (SNR) of the set of signals. The range of SNR in the set $V$ was kept from 1 to 50. Figure \ref{fig:toy}(a) shows some sample sinusoidal waves in the set $V$ with different SNR. Figure \ref{fig:toy}(b) shows the variation of \textit{ApEn} with varying SNR and radius. As expected, we can see that the higher the SNR (lesser noise), the lower the value of \textit{ApEn} gets for any value of $r$.

%\begin{figure}[b]
%	\centering
%	\includegraphics[width=1.0\columnwidth]{images/toy/ApEn_toy.png}
%	\caption{ }
%	\label{fig:ApEn_toy}
%	%	\vspace{-2pt}
%\end{figure}
%In order ot show the dependency of ApEn features on data predictability, we generate a toy data set with sine waves with different amount of white gaussian noise added. Figure \ref{fig:apen_toy} shows some of the sample sine waves with different Signal-to-Noise (SNR) ratios. We can see that the lower the SNR, the more irregular and unpredictable the wave becomes. Therefore a lower SNR would result in higher entropy which is proven by the graph of ApEn vs SNR given in Figure \ref{fig:apen_toy}.

%\begin{figure}[t]
%	\centering
%	\includegraphics[width=1.0\columnwidth]{images/ApEn_toy_example_new.png}
%	\caption{A toy example for showing how approximate entropy (ApEn) varies with noise in a signal. The three plots on the left show sample sine waveforms and their corresponding signal-to-noise (SNR) ratios. The plot on the right shows how the value of ApEn varies for different values of radius used with respect to SNR}
%	\label{fig:apen_toy}
%	\vspace{-2pt}
%\end{figure}

\noindent\emph{\textbf{Cross Approximate Entropy:}} 
Cross approximate entropy \textit{`XApEn'} is a measure of asynchrony between two time series. For two given time series $[T,S] \in \Re^N$, the embedding vectors are defined as $x_1(i) = [T_i,T_{i+\tau},\ldots,T_{i+(m-1)\tau}]$ and $x_2(i) = [S_i,S_{i+\tau},\ldots,S_{i+(m-1)\tau}]$, for $ 1\leq i\leq N-(m-1)\tau$. The frequency of repeatable patterns $C^m_i(r)(T||S)$ for the embedding vectors $x_1(i)$ and $x_2(i)$ is then calculated by
\begin{equation}
	C^m_i(r)(T||S) = \frac{1}{N-(m-1)\tau}\sum_{j}H(r-dist(x_1(i),x_2(j)))
\end{equation}
$\Omega^m(r)$ is then calculated using
\begin{equation}
	\Omega^m(r) = \frac{1}{N-(m-1)\tau}\sum_{i=1}^{N-(m-1)\tau}ln(C^m_i(r)(T||S))
\end{equation}
This is then used to finally calculate the cross approximate entropy between the two time series by $XApEn(m,r,\tau) = \Omega^m(r)(T||S) -\Omega^{m+1}(r)(T||S).$
%\vspace{-5pt}

Similar to \textit{ApEn}, we generate a set of sinusoids $W$ to show the variation of \textit{XApEn} for varying synchrony between different signals. The set $W$ consists of sinusoids with the same SNR but with phase varying from 0 to $\pi$. Figure \ref{fig:toy}(c) shows some sample of sinusoids in this set. Figure \ref{fig:toy}(d) shows how the value of \textit{XApEn} varies when the phase difference between the signals varies. We can see that the value of \textit{XApEn} reaches a max at about $0.5\pi$ and then reduces back to 0 at $\pi$ phase difference. It is important to note that two sinusoids with a phase difference of $\pi$ are completely out of phase but in perfect synchrony. This is because if one increases the other decreases with the same rate. This should result in a very low \textit{XApEn} value which we observe in Figure \ref{fig:toy}(d) as well.

%\noindent{\emph{\textbf{Step 4: Feature selection and skill classification}}}: Since all DCT and DFT frequency coefficients may not be relevant for skill assessment, we perform feature selection to determine a subset of skill.

\begin{figure}[t]
	\centering
	\includegraphics[width=1.0\columnwidth]{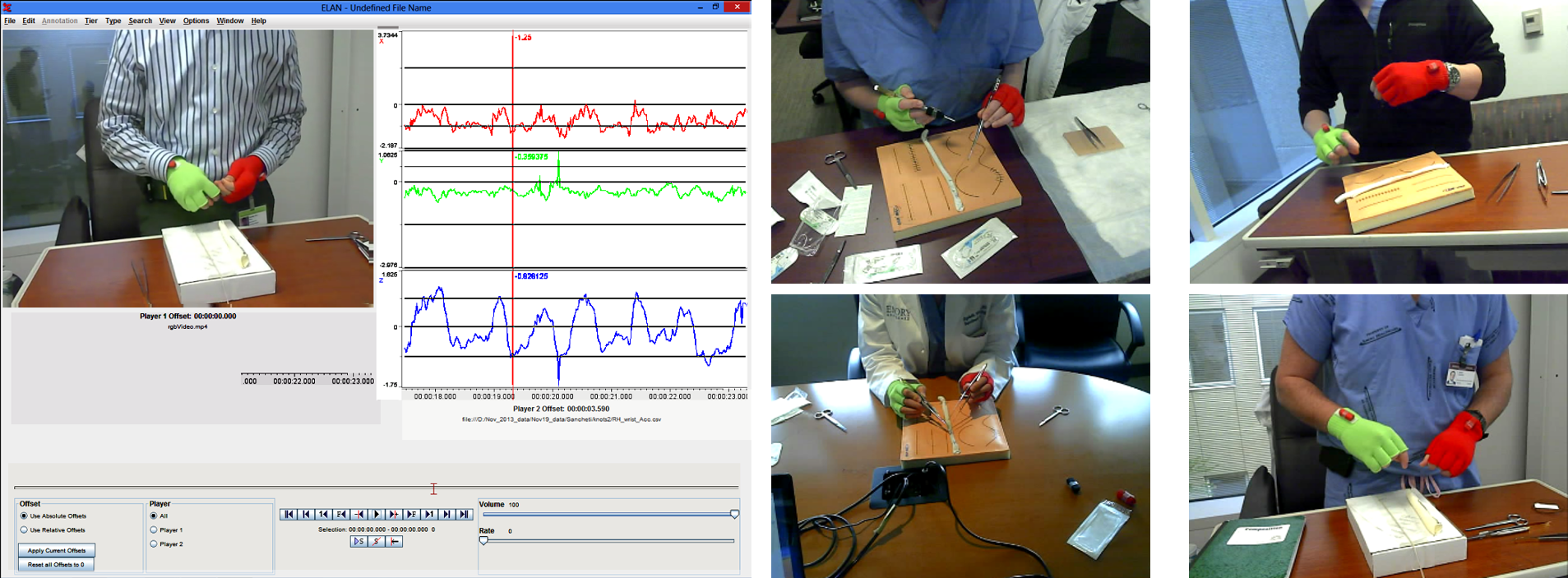}
	\caption{Image on left shows a screenshot from ELAN software for synchronization of video and accelerometer data. Middle column and right most column shows sample frames for suturing and knot tying, respectively. The accelerometers can also be seen placed on the wrists and the needle-holder}
	\label{fig:elan}
%	\vspace{-2pt}
\end{figure}

Surgical motions in suturing and knot tying tasks are inherently repetitive in nature. The repetitiveness of motion can be encoded using frequency features. However, frequency features would not be able to capture the sudden movements or jerks in motion that define the competitiveness of a surgeon. They do not quantify the orderliness or predictability of patterns. On the other hand, approximate entropy represents the likelihood of occurrence of similar patterns of observations. A time series containing many repetitive patterns has lower approximate entropy and is more predictable.Therefore, using \textit{`ApEn'} features can potentially capture repetitiveness along with more finer details crucial for skills assessment. Moreover, in surgical motions, it is also important for surgeons to move their hands and tools in a smooth motion together. We think that \textit{`XApEn'} features can potentially capture information on how synchronized the surgeon's hands and tools are with each other. We use both the entropy based features described above to encode surgical motion predictability for our analysis. 

%\vspace{-3pt}
\section{Experimental Evaluation}
%\vspace{-5pt}

\subsection{Data Set}
%\vspace{-6pt}
Our data set consists of video and accelerometer data for evaluating the performance of proposed and previous state-of-the-art features for skill assessment. We use the surgical skills dataset from \cite{zia2015automated} for direct comparisons.  This dataset had 18 participants. We augmented this dataset with additional 23 participants to a total of 41 participants consisting of surgical residents and nurse practitioners, essentially doubling the data set from previous studies. In this data set, each participant undertook two instances each of suturing and knot tying tasks. For each instance, video data was captured at 30 frames per second at a resolution of $640\times480$ using a standard RGB camera. We collected 4000 and 1000 frames for each trial of suturing and knot tying, respectively. Each video was captured in different lighting conditions and from varying camera angles to make the data set invariant to lighting and viewing angle. Figure \ref{fig:elan} shows some sample frames from the videos. Due to acquisition errors, some videos had to be excluded from the data set resulting in 74 videos for each surgical task.

The acceleration data was captured using Axivity sensors. Two accelerometers were used for each surgical task. For knot tying, one accelerometer was attached to each hand wrist whereas for suturing, one accelerometer was attached to the dominant hand wrist and one to the needle-holder. The data captured consisted of $x$, $y$ and $z$ acceleration values resulting in a 3-dimensional time series for each accelerometer. At the start of each instance, all participants were asked to rapidly shake the hands/instruments with the accelerometers to get the synchronization waveform that is used to align the starting point of acceleration data with the video using the ELAN software (a snapshot shown in figure \ref{fig:elan}). The accelerometer data had some additional noise as the accelerometers were not being attached properly, resulting in unwanted jerks. For some cases, the accelerometer even fell off during a session and had to be reattached. All such cases were removed from the data set resulting in a final 54 acceleration data recordings for knot tying and 62 for suturing. The average length of acceleration time series data was 8434 for suturing and 1919 for knot tying. A complete class distribution for video and accelerometer data is given in Table \ref{class_distribution}.
%\vspace{-5pt}
% \usepackage{multirow}
% \usepackage{graphicx}
\begin{table}[t]
	\centering
	\caption{Skill class distribution. Each cell contains two values $V:A$, where `$V$' = No. of participants for video data, `$A$' = No. of participants for acceleration data.}
	\label{class_distribution}
	\resizebox{\textwidth}{!}{%
		\begin{tabular}{|c|c|c|c|c|c|c|c|c|c|}
			\hline
			\multirow{2}{*}{}     & \multicolumn{5}{c|}{\textbf{Suturing}}                              & \multicolumn{4}{c|}{\textbf{Knot Tying}}              \\ \cline{2-10} 
			& \textbf{RT} & \textbf{TM} & \textbf{IH} & \textbf{SH} & \textbf{FO} & \textbf{TM} & \textbf{SH} & \textbf{FO} & \textbf{OP} \\ \hline
			\textbf{Beginner}     & 38 : 28     & 46 : 34     & 47 : 35     & 47 : 35     & 45 : 33     & 27 : 18     & 27 : 19     & 22 : 15     & 23 : 15     \\ \hline
			\textbf{Intermediate} & 22 : 20     & 15 : 15     & 13 : 13     & 17 : 17     & 18 : 18     & 22 : 17     & 28 : 21     & 28 : 22     & 28 : 22     \\ \hline
			\textbf{Expert}       & 14 : 14     & 13 : 13     & 14 : 14     & 10 : 10     & 11 : 11     & 25 : 19     & 19 : 14     & 24 : 17     & 23 : 17     \\ \hline

		\end{tabular}%
	}
\end{table}

%\vspace{-6pt}
\subsection{Parameter Selection}
\vspace{-6pt}
Both the entropy based features proposed in this paper were evaluated on our data set. In order to compare the performance of these features, we also evaluate previous state-of-the-art methods such as SMT \cite{sharmavideo} and DCT/DFT \cite{zia2015automated} in the same experimental setup. Traditional methods such as HMM, BoW and A-BoW were reported to perform poorly as compared to SMT and DCT/DFT features in \cite{zia2015automated} and hence were excluded from the experiments.

We used $K \in [2,3,4,5,6,7,8,9,10,12,14,16,18,20]$ for $k$-means clustering to learn motion classes (the number of time series dimensions used) for analysis of video data. The accelerometer data, however, did not have this dependency with a 6-dimensional time series (concatenation of 3-dimensional time series from two accelerometers used) for all evaluations. SMT and frequency based features (DCT and DFT) were implemented as presented in \cite{sharmavideo,zia2015automated} for both modalities. As described in the previous section, entropy based features are dependent on some parameters which need to be specified. These are mainly the embedding dimension ($m$) and the radius ($r$). In order to be able to differentiate time series data on the basis of regularity, radius ($r$) can have value between 0.1 to 0.25 times the standard deviation of the time series, whereas $m=1$ and $m=2$ both work equally well \cite{pincus1991approximate}.

For \textit{ApEn}, the approximate entropy for each dimension of the time series is calculated for values of $r = [0.1,0.13,0.16,0.19,0.22,0.25]$ resulting in a feature vector $\theta_{ApEn} \in \Re^{6K}$, where $K$ is the dimension of time series used (6 for acceleration data but variable for video data). However, for \textit{XApEn}, we use the same values of $r$ for accelerometer data but only use $r=0.2$ for videos. This was done since it was observed that the value of \textit{XApEn} did not vary much for different values of $r$ for videos. Moreover, the computation time for \textit{XApEn} also increases significantly with increasing dimensionality of time series as is the case for videos. We obtain a final feature vector for cross entropy $\theta_{XApEn} \in \Re^{\frac{RK(K-1)}{2}}$, where $R$ denotes the number of radius values used in evaluation. We also check the performance of fusing \textit{ApEn} and \textit{XApEn} before classification by concatenation resulting in a feature vector $\theta_{ApEn+XApEn} \in \Re^{\frac{RK^2+K(12-R)}{2}}$.

The value of $m$ is set as 1 for all evaluations. For fair comparisons with previously proposed techniques, we use similar classification methodology and adopt Leave-one-out cross validation (LOOCV) and use a Nearest Neighbor (NN) classifier after selecting features using SFFS. 

%Moreover, in this paper, we also do a performance analysis of top performing features by varying the amount of training data used. For that, we use $P\%$  of data for training and test on the rest. The classification accuracy is averaged over 100 repetitions of each trial. We vary the value of $P$ from 10 till 90.

%\begin{figure}[h]
%	\centering
%	\includegraphics[width=1.0\columnwidth]{KnotTying_average_results.png}
%	\caption{}
%	\label{fig:KnotTying_average_results}
%	\vspace{-2pt}
%\end{figure}
%Creative Intel Perceptual camera~\cite{intelcam}. The Creative camera is both economical and easy to use and setup. It also captures depth information which we plan to use in our future work. 

%Figure 2

%TAble 1

%TABLE 1

%\begin{figure}[h]
%	\centering
%	\includegraphics[width=1.0\columnwidth]{KnotTying_kfold.png}
%	\caption{}
%	\label{fig:KnotTying_kfold}
%	\vspace{-2pt}
%\end{figure}

%\begin{figure}[h]
%	\centering
%	\includegraphics[width=1.0\columnwidth]{Suturing_kfold.png}
%	\caption{}
%	\label{fig:Suturing_kfold}
%	\vspace{-2pt}
%\end{figure}

%\begin{figure}[b]
%	\centering
%	\includegraphics[width=1.0\columnwidth]{images/holdout.png}
%	\caption{Average classification accuracies versus percentage of data used for video data}
%	\label{fig:holdout}
%	\vspace{-2pt}
%\end{figure}

\begin{figure}[b]
	\centering
	\includegraphics[width=1.0\columnwidth]{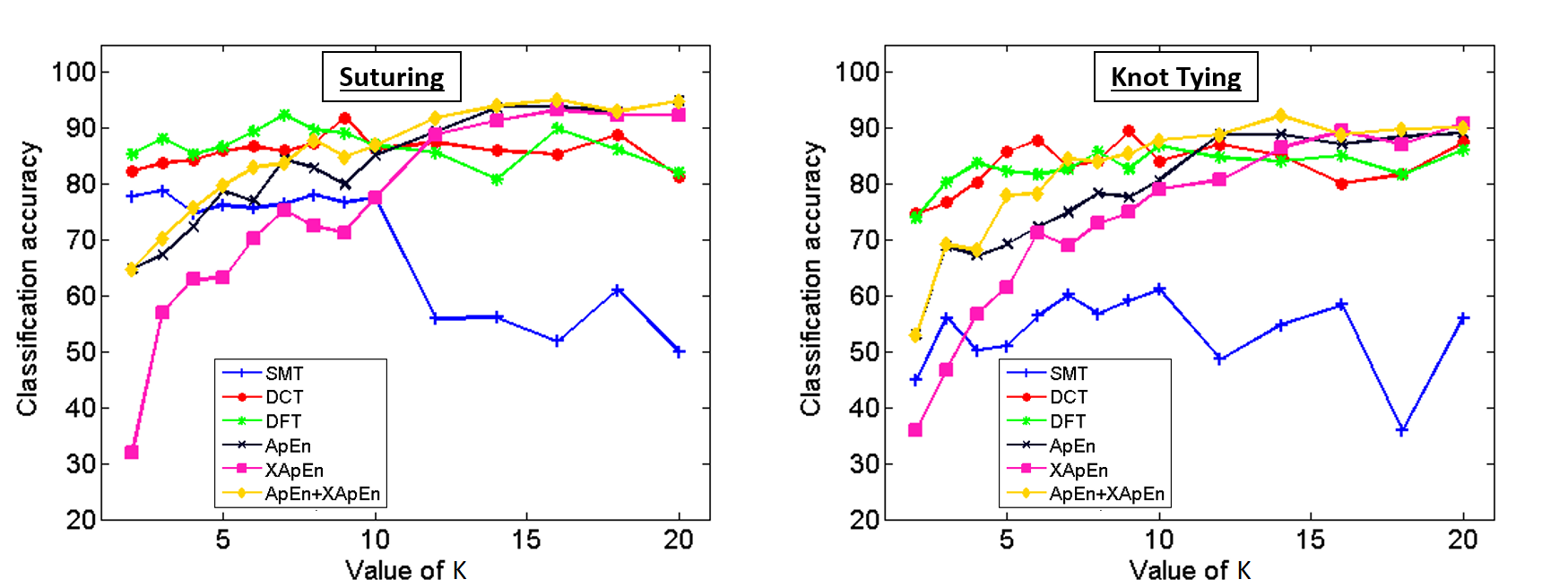}
	\caption{Average classification accuracy versus value of $K$ used (number of dimensions of time series) for video data only. (Best viewed in color)}
	\label{fig:average_results}
%	\vspace{-2pt}
\end{figure}

%\vspace{-10pt}
\section{Results and Discussion}
\vspace{-6pt}
The features described in the previous section were evaluated on video and accelerometer data for suturing and knot tying tasks for all applicable OSATS criteria. For video, we calculate the average classification accuracy over all OSATS criteria of different features for all the values of $K$ in order to find the optimum number of clusters to use for each feature type. The average accuracy $\hat{A}$ is calculated using $\hat{A} = \frac{1}{O}\sum\limits_{OSATS} A_K$, where $A_K$ is the accuracy using $K$ clusters for a specific OSATS criteria, and $O$ represents the total number of applicable OSATS criteria for that task. Figure \ref{fig:average_results} shows the comparison of different features for suturing and knot tying tasks using video data. We can see that entropy based features are able to achieve the highest average accuracy for both suturing and knot tying tasks using combined ApEn and XApEn features. 

\begin{figure}[t]
	\centering
	\includegraphics[width=1.0\columnwidth]{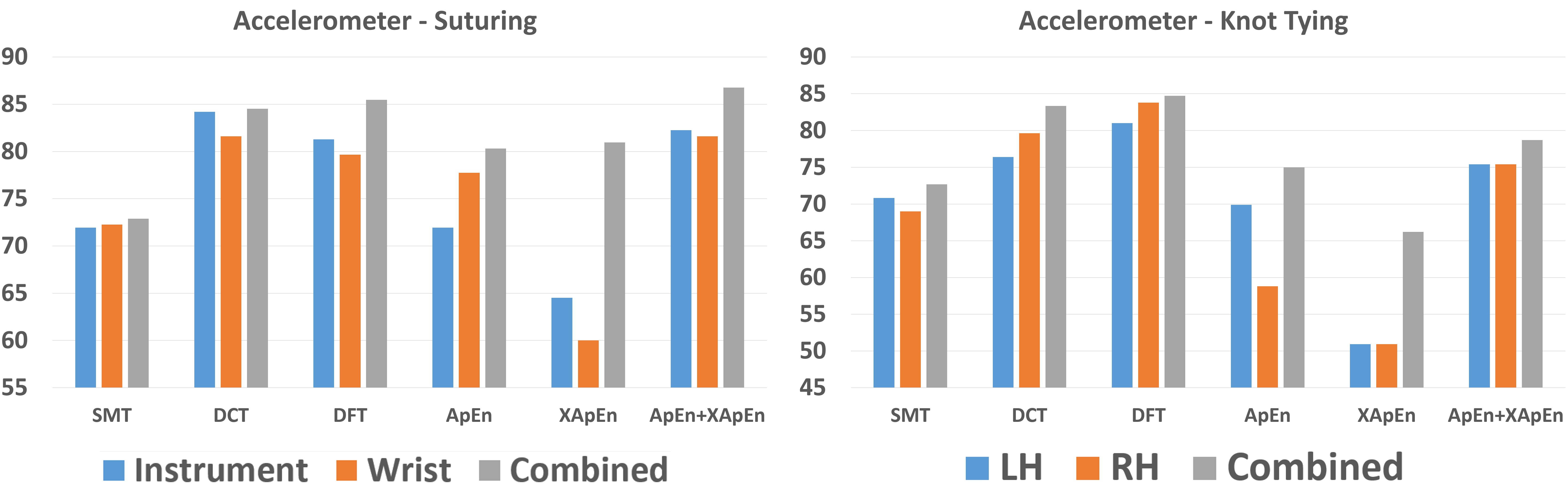}
	\caption{Average classification results for accelerometer data using individual and combination of the two accelerometers attached. (Best viewed in color) }
	\label{fig:Results_Acc}
	%	\vspace{-2pt}
\end{figure}

For accelerometer data, we evaluate the different features for both the accelerometers attached for each task; wrist and needle-holder for suturing and hand wrists for knot tying. Figure \ref{fig:Results_Acc} shows the average classification results achieved. Its evident from Figure \ref{fig:Results_Acc} that the combination of data from both accelerometers performs better than individual accelerometers for both tasks and all feature types. However, it is interesting to see a difference in accuracy achieved using accelerometers attached on different positions for skill assessment. This analysis can show which motions are more skill relevant for the two tasks at hand and can potentially be used to give better feedback to surgeons on how to improve their performance.

\begin{table}[b]
	\centering
	\caption{Highest average classification accuracies with standard deviations for different techniques using multi-modality data. For video data, $K$ corresponding to highest accuracy is also shown.}
	\label{results}
	\resizebox{\textwidth}{!}{
		\begin{tabular}{|c|c|c|c|c|}
			\hline
			\multirow{2}{*}{} & \multicolumn{2}{c|}{Video}                                  & \multicolumn{2}{c|}{Accelerometer}            \\ \cline{2-5} 
			& Suturing                     & Knot Tying                   & Suturing              & Knot Tying            \\ \hline
			SMT               & 78.9 $\pm$ 5.7 (K=3)             & 61.1 $\pm$ 2.3 (K=10)            & 72.9 $\pm$ 4.5            & 72.7 $\pm$ 5.3            \\ \hline
			DCT               & 91.9 $\pm$ 3.4 (K=9)             & 89.5 $\pm$ 2.8 (K=9)             & 84.5 $\pm$ 4.9            & 83.3 $\pm$ 2.1            \\ \hline
			DFT               & 92.4 $\pm$ 3.7 (K=7)             & 86.8 $\pm$ 2.8 (K=10)            & 85.5 $\pm$ 3.0            & \textbf{84.7 $\pm$ 4.1} \\ \hline
			ApEn              & 93.7 $\pm$ 2.2 (K=20)            & 89.2 $\pm$ 5.3 (K=20)            & 80.3 $\pm$ 2.1            & 75.0 $\pm$ 6.5            \\ \hline
			XApEn             & 91.4 $\pm$ 3.0 (K=16)            & 90.9 $\pm$ 4.3 (K=20)            & 81.0 $\pm$ 4.0            & 66.2 $\pm$ 4.1            \\ \hline
			ApEn+XApEn        & \textbf{95.1 $\pm$ 3.1 (K=16)} & \textbf{92.2 $\pm$ 3.0 (K=14)} & \textbf{86.8 $\pm$ 4.5} & 78.7 $\pm$ 5.8            \\ \hline
		\end{tabular}
	}
\end{table}

The highest average accuracy and the corresponding standard deviations achieved for different techniques are given in Table \ref{results}. We can see that entropy based features perform better for video data as compared to state-of-the-art techniques. For accelerometer data, entropy based features perform better for suturing but not so well for knot tying. The reasons for this is mainly because entropy based features are dependent on the dimension of the time series used (can also be seen in Figure \ref{fig:average_results} for increasing values of $K$); the higher the dimension of time series being evaluated, the more information is captured especially for cross entropy (\textit{XApEn}). In case of accelerometer data, we only have 3-axis acceleration values so entropy based features cannot capture enough information. However, it is interesting to see that entropy based features still perform better for suturing task. 

\begin{figure}[t]
	\centering
	\includegraphics[width=1.0\columnwidth]{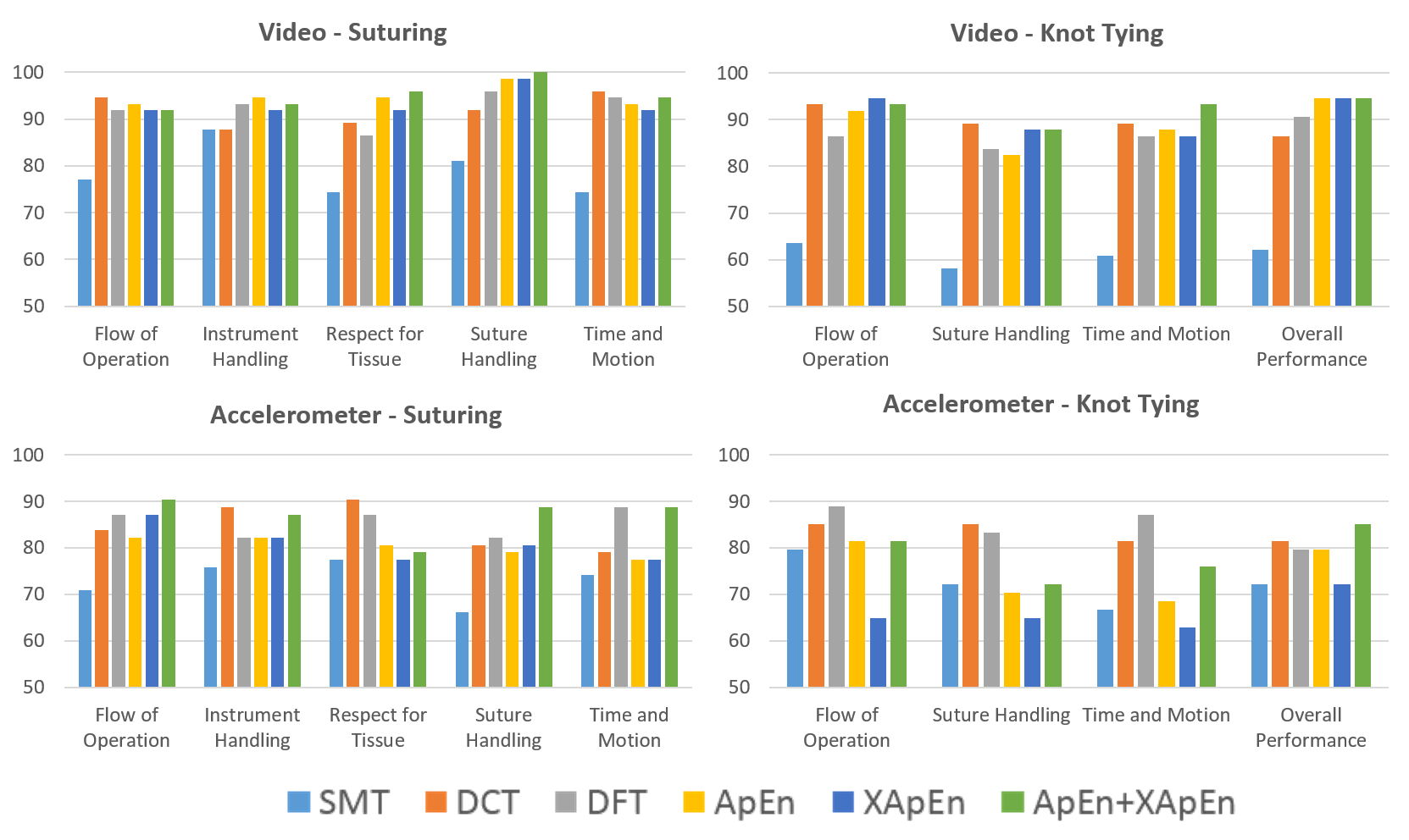}
	\caption{Individual OSATS criteria results for video and accelerometer data. For each feature, the optimal value of $K$ (as indicated in Table \ref{results}) was used. (Best viewed in color)}
	\label{fig:Results_MaxAverage}
	%	\vspace{-2pt}
\end{figure}

Figure \ref{fig:Results_MaxAverage} shows the results for individual OSATS criteria by using the optimal $K$ for each feature type (as indicated in Table \ref{results}). Comparing the two modalities, we see that all the techniques perform better for video as compared to accelerometer. This can be explained by the fact that accelerometers only capture the hands/needle-holder 3-D acceleration data whereas videos can be used to extract all motions~(both hands, instruments etc.). 

Comparing results for individual modalities shows us that using video data performs much better than accelerometer for all feature types. However, it is possible that a fusion of features from video and accelerometer data performs better compared to individual modalities. Therefore, we adopt an early fusion scheme and run our analysis for frequency(DCT and DFT) and best performing entropy features (ApEn+XApEn). The features are fused via concatenation. Since some of the accelerometer data had to be excluded (as described in Section 4), we only use videos for which the corresponding accelerometer data is available i.e 54 for knot tying and 62 for suturing. Tables \ref{results-modalities-suturing} and \ref{results-modalities-knot tying} show the average accuracies (over all OSATS criteria) with standard deviations using different modalities for suturing and knot tying, respectively. We can see that combining video and accelerometer data deteriorates performance for DCT and DFT features as compared to video. For ApEn+XApEn, the performance improves for knot tying but has a slight decrease as compared to video for suturing. Overall, the highest performance is achieved using ApEn+XApEn features for each task (shown in bold). 

\begin{figure}[t]
	\centering
	\includegraphics[width=1.0\columnwidth]{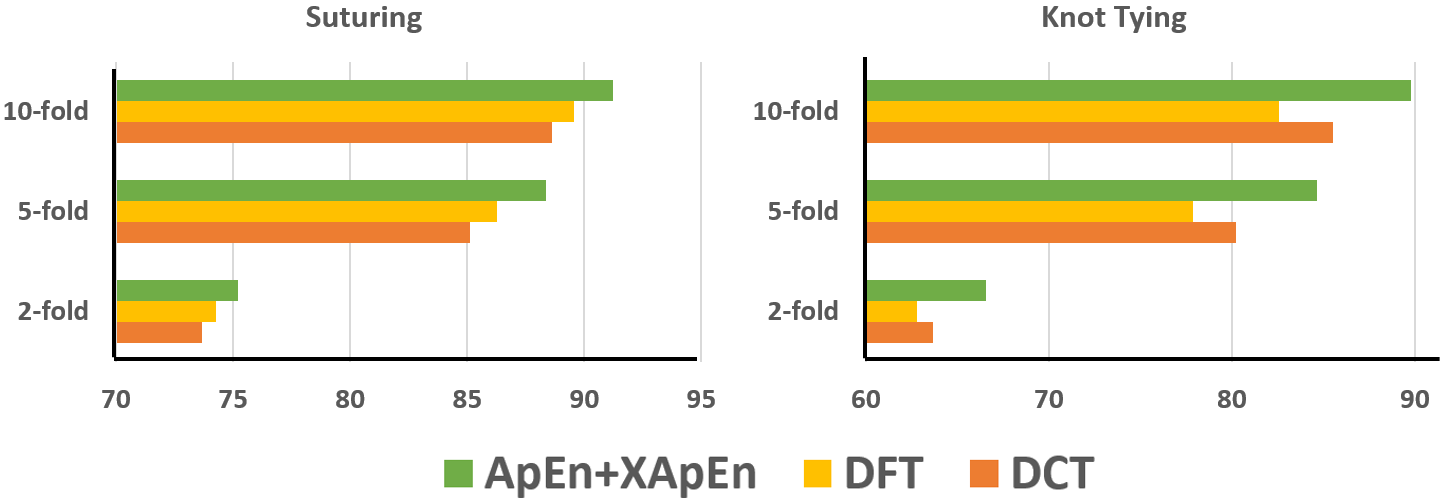}
	\caption{Average classification accuracies for different cross validation schemes by using highest performing modality. For DCT and DFT, video data was used for both tasks. Whereas, for ApEn+XApEn, video data was used for suturing and video+accelerometer data was used for knot tying (see Tables \ref{results-modalities-suturing} and \ref{results-modalities-knot tying} for best performances). (Best viewed in color)}
	\label{fig:Results_Cross_Validation}
	%	\vspace{-2pt}
\end{figure}

In order to check the robustness of different features, we perform another experiment by using harder cross validation schemes of 2,5 and 10 fold. We again compare ApEn+XApEn with DCT and DFT. For this analysis, the best performing modality for each feature being compared was used. Therefore, we use video data with DCT and DFT for both tasks, whereas, we use video data for suturing and video+accelerometer data for knot tying with ApEn+XApEn (refer to Tables \ref{results-modalities-suturing} and \ref{results-modalities-knot tying}). Figure \ref{fig:Results_Cross_Validation} shows the average accuracies over all OSATS criteria for the different cross validation schemes used. One can see that the proposed ApEn+XApEn features outperform frequency based features for all cross validation schemes. This shows that the proposed entropy based features are also robust to the amount of training data available as compared to frequency features. 

Although the previously proposed frequency features perform reasonably well (especially for accelerometer data), we think that they perform well on repetitive surgical tasks like suturing and knot tying. We believe that the proposed entropy based features would perform better in other surgical procedures as well since they try to capture the irregularity in motion instead of just the repetitiveness.
%AN IMPORTANT POINT TO NOTE IS THAT WE HAVE LOWER ACCURACY FOR KNOT TYING AS COMPARED TO SUTURING WITH ALL FEATURE TYPES. THIS NEEDS SOME EXPLANATION, NOT SURE HOW TO BEST PRESENT THIS.

%\begin{table}[b]
%	\centering
%	\caption{Average accuracies with standard deviations using different data modalities for frequency and best performing entropy based feature. Highest performance across all modalities and feature types for each task is shown in bold}
%	\label{results-video-accelerometer}
%	\resizebox{\textwidth}{!}{
%	\begin{tabular}{|c|c|c|c|c|c|c|}
%		\hline
%		\multirow{2}{*}{} & \multicolumn{3}{c|}{Suturing}                   & \multicolumn{3}{c|}{Knot Tying}                 \\ \cline{2-7} 
%		& Video     & Accelerometer & Video+Accelerometer & Video     & Accelerometer & Video+Accelerometer \\ \hline
%		DCT               & 90.6 $\pm$ 3.1 & 84.5 $\pm$ 4.9     & 86.8 $\pm$ 7.7           & 91.7 $\pm$ 6.1 & 83.3 $\pm$ 2.1     & 83.8 $\pm$ 4.9           \\ \hline
%		DFT               & 87.1 $\pm$ 1.1 & 85.5 $\pm$ 3.0     & 86.1 $\pm$ 2.1           & 86.1 $\pm$ 1.9 & 84.7 $\pm$ 4.1     & 81.0 $\pm$ 5.5           \\ \hline
%		ApEn+XApEn        & \textbf{93.9 $\pm$ 3.7} & 86.8 $\pm$ 4.5     & 93.2 $\pm$ 6.6           & 90.3 $\pm$ 3.1 & 78.7 $\pm$ 5.8     & \textbf{94.0 $\pm$ 2.8}           \\ \hline
%	\end{tabular}
%	}
%\end{table}

\begin{table}[t]
	\centering
	\caption{Average accuracies with standard deviations for corresponding feature types using different data modalities for suturing task. Highest performance across all modalities and feature types is shown in bold}
	\label{results-modalities-suturing}
%	\resizebox{\textwidth}{!}{
	\begin{tabular}{|c|c|c|c|}
		\hline
		& Video                     & Accelerometer  & Video+Accelerometer \\ \hline
		DCT        & 90.6 $\pm$ 3.1            & 84.5 $\pm$ 4.9 & 86.8 $\pm$ 7.7      \\ \hline
		DFT        & 87.1 $\pm$ 1.1            & 85.5 $\pm$ 3.0 & 86.1 $\pm$ 2.1      \\ \hline
		ApEn+XApEn & \textbf{93.9 $\pm$ 3.7} & 86.8 $\pm$ 4.5 & 93.2 $\pm$ 6.6      \\ \hline
	\end{tabular}
%	}
\end{table}

\begin{table}[t]
	\centering
	\caption{Average accuracies with standard deviations for corresponding feature types using different data modalities for knot tying task. Highest performance across all modalities and feature types is shown in bold}
	\label{results-modalities-knot tying}
	\begin{tabular}{|c|c|c|c|}
		\hline
		& Video          & Accelerometer  & Video+Accelerometer       \\ \hline
		DCT        & 91.7 $\pm$ 6.1 & 83.3 $\pm$ 2.1 & 83.8 $\pm$ 4.9            \\ \hline
		DFT        & 86.1 $\pm$ 1.9 & 84.7 $\pm$ 4.1 & 81.0 $\pm$ 5.5            \\ \hline
		ApEn+XApEn & 90.3 $\pm$ 3.1 & 78.7 $\pm$ 5.8 & \textbf{94.0 $\pm$ 2.8} \\ \hline
	\end{tabular}
\end{table}

%\vspace{-10pt}
\section{Conclusion}
%\vspace{-6pt}
We presented a comparison of the proposed entropy based features for assessment of surgical skills using video and accelerometer data with previous state-of-the-art. Overall, our analysis showed that videos are better for extracting skill relevant information as compared to accelerometer. However, a fusion of video and accelerometer features can improve on performance.  Also, the proposed combination of \textit{ApEn} and \textit{XApEn} outperforms state-of-the-art features.

Having an automated system for surgical skills assessment can significantly improve the quality of surgical training. It would allow the surgical trainees to practice their basic skills a lot more with valuable feedback.  Moreover, such a system could also save time for expert surgeons that is spent on trainee assessment. 
\bibliographystyle{splncs}
\bibliography{IPCAI}   % name your BibTeX data base

\end{document}